\documentclass[a4paper, 10pt, conference]{ieeeconf}      
 \pdfoutput=1
                                                                                                         
\usepackage{FG2020}

\FGfinalcopy 

\IEEEoverridecommandlockouts                              
\usepackage{graphics} 
\usepackage{epsfig} 
\usepackage{mathptmx} 
\usepackage{times} 
\usepackage{amsmath} 
\usepackage{amssymb}  
\usepackage[utf8]{inputenc}
\newcommand{\bftab}{\fontseries{b}\selectfont}

\def\FGPaperID{93} 

\title{\LARGE \bf Neural Sign Language Translation by Learning Tokenization }

\author{\parbox{16cm}{\centering
    {\large Alptekin Orbay and Lale Akarun }\\
    {\normalsize
    Faculty of Computer Science, Bo\u{g}azi\c{c}i University, Turkey\\}}
    \thanks{This work was supported by the Scientific and Technological Research Council of Turkey (TÜBİTAK) Project \#117E059. }
}

\begin{document}

\ifFGfinal
\thispagestyle{empty}
\pagestyle{empty}
\else
\author{Anonymous FG2020 submission\\ Paper ID \FGPaperID \\}
\pagestyle{plain}
\fi
\maketitle

\begin{abstract}
    Sign Language Translation has attained considerable success recently, raising hopes for improved communication with the Deaf. A pre-processing step called tokenization improves the success of translations. Tokens can be learned from sign videos if supervised data is available. However,  data annotation at the gloss level is costly, and annotated data is scarce. The paper utilizes Adversarial, Multitask, Transfer Learning to search for semi-supervised tokenization approaches without burden of additional labeling. It provides extensive experiments to compare all the methods in different settings to conduct a deeper analysis. In the case of no additional target annotation besides sentences, the proposed methodology attains 13.25 BLUE-4 and 36.28 ROUGE scores which improves the current state-of-the-art by 4 points in BLUE-4 and 5 points in ROUGE.
\end{abstract}

\section{INTRODUCTION}

Sign Languages are the primary languages of the Deaf. Sign Language (SL) has its own linguistic structures like spoken and written languages. Interpreting the sign language to ordinary people definitely alleviates the burden for the deaf who suffer from the lack of communication in daily life. Automatic Sign Language Recognition has aimed to identify signs or utterances for communicating with the Deaf \cite{koller16:deephand},\cite{camgoz2017subunets}. However, translation from sign language to spoken language is the ultimate goal. To achieve this, proposed approaches ought to cover linguistic properties as well as visual properties of sign languages. Little work has been done in this field until recently \cite{camgoz2018neural}.

\par Deep Learning (DL) \cite{dl} has opened new opportunities to deal with various tasks of Computer Vision (CV) and Natural Language Processing (NLP). Deeper models \cite{alexnet} made a dramatic leap in image classification and it is indicated that RNNs are good at temporal modelling and generating sequences in \cite{rnn}. Thanks to those achievements, DL has dominated machine translation \cite{bahdanau}, \cite{luong-etal-2015-effective}, \cite{transformer} and replaced feature engineering by enabling strong feature extractors \cite{inception},  \cite{cao2018openpose}, \cite{I3D} in both images and videos. All of these show promise for significant improvements in sign language translation. 
	\par In this paper, we study specific challenges for sign language translation. The first challenge is that there is more data for SL tasks like hand shape recognition, but limited data for the translation task. We believe that this limits the success of end-to-end learning, which performs very well in other domains. Collecting data is laborious and costly as annotation requires translators skilled in sign languages. Annotators have to provide  glosses, which are intermediate representations between signs and words.  To deal with these challenges, we investigate methods to compensate the absence of data. Camgoz et al. \cite{camgoz2018neural} formalized NSLT  with two stages, an implicit tokenization layer and a sequence-to-sequence model. In \cite{camgoz2018neural}, the focus has been on end-to end learning and optimization of tokenization has been left as future work. The tokenization layer aims to extract good representations from videos and converts them into tokens. The sequence-to-sequence translation seeks for an optimal mapping from those tokens to sentences by learning a language model. The latter is an active area of research whereas tokenization, or how videos should be represented to be fed to sequence-to-sequence modeling has attracted less attention so far. Gloss-to-text translation seems to be more accurate than sign-to-text translation. However, SL resources with gloss annotations are very scarce. We  apply and  compare different  approaches successful in other domains for  unsupervised or semi-supervised learning of tokenization. To prevent additional labeling for glosses, we look for ways to  increase the quality of sign-to-text translation. Our contributions may be summarized as:
	\begin{itemize}
	\item We have illustrated the low performance of the  end-to-end scheme in the absence of sufficient data.
	\item We have applied Adversarial, Multitask, and Transfer Learning to leverage additional supervision.
	\item We have experimented with 3D-CNN and 2D-CNN models specialized on hand shape and human pose to search effective tokenization methods.
	\item We have used  datasets in different sign languages with multiple tasks and proposed a target domain independent frame-level tokenization method exceeding the current state-of-the-art method dramatically.
	\end{itemize}

\section{Related Work}
	Sign Language Translation (SLT) has recently begun to attract the interest of researchers \cite{camgoz2018neural}, \cite{chinesse},\cite{korean}.  The progress of this domain is restricted by the lack of sufficient data as special effort and knowledge is needed to collect and label sign language. The majority of large datasets have been collected  either for educational purposes or for studying the linguistic properties \cite{danish}, \cite{nz}.  As a consequence, the data is often weakly labeled and requires additional effort for learning.  Other datasets have been collected for human computer interaction purposes and consist of isolated signs \cite{chinesse}, \cite{bosphorus} and are not useful for sign language translation. 
	 \par The weak labeling challenge has been addressed in recent papers \cite{Pfister}, \cite{Buehler} ,\cite{hellen}. The new techniques provide large scale learning in the absence of strong annotation. TV broadcasts include abundant data, but they are often not annotated. After Forster et al. released RWTH-PHOENIX-Weather 2012 \cite{forster1}, its extended version RWTH-PHOENIX-Weather 2014 \cite{forster2}, gloss and sentence level annotations has been made available. 
	\par DL has begun to dominate sign language research  as well as other domains. CNNs have become powerful approaches for hand shape classification \cite{koller16:deephand} and RNNs have been used for temporal modeling in sign language videos \cite{camgoz2017subunets}. The idea of sequence to sequence learning \cite{seq2seq} is one of the most important advances in DL. It proposes effective encoder-decoder architectures  to find an optimal mapping from one sequence to another. However, a serious problem is that  various length sequences are converted into fixed-sized vectors. Regarding this, RNN-based attentional seq2seq architectures  \cite{bahdanau}, \cite{luong-etal-2015-effective} are proposed and a new self-attentional network \cite{transformer} is introduced. These successful attempts have paved the way for NSLT to handle long video sequences and generate qualified translations.
\par  The first proposed tokenization method by Camgoz et al. \cite{camgoz2018neural} uses a 2D-CNN and a sequence-to-sequence model combined in tandem. They are trained jointly with translation data in an end-to-end scheme. In the tokenization layer, the CNN extracts features from each image and features are concatenated in time to feed the sequence-to-sequence model. It is assumed that the CNN can learn good representations from images trained on the sentence level annotation. However, the validity of this assumption is not tested.  Effective methods for extracting keypoints of body parts from 2D images have been developed \cite{cao2018openpose} and methods based on extracting features from these keypints have been proposed\cite{korean},\cite{open-chinese}.  These  key point detectors are robust to different backgrounds and camera setups within limits. Sign language videos suffer from them along with motion blur. However, there is no clear evidence to show that noisy hand key points are capable of representing the hand shape. We know that subtle changes in hand shapes can lead to completely different meanings. Apart from translation research, there have been studies to find optimal representations from videos in sign language research. 2D-CNNs \cite{camgoz2017subunets} are used to extract frame level features and 3D-CNNs \cite{3d1},\cite{3d2} are considered as video level tokenizers. We apply and compare available methods discussed above for translation in the following sections. 

\section{Learning Tokenization}
	It is desired to cover linguistic and visual properties of SL to attain better translations. As seen in Fig. 1, NSLT system can be divided into two parts, Neural Machine Translation and Tokenization. For linguistic perspective, SLs are highly dependent on the context of communications and a NSLT  must handle this problem by giving attention on short term information and long term dependencies. Attentional Encoder-Decoder mechanisms are designed to address this problem and have achieved considerable but limited success. Strong tokenizations may handle visual properties of SL and provide good representations either at the frame level or at the video level. That also contributes to solving the linguistic challenges. 
	Therefore, we discuss the alternative tokenization approaches leaving Neural Machine Translation as future work.
	\begin{figure}[t]
\centering
\includegraphics[width=8cm]{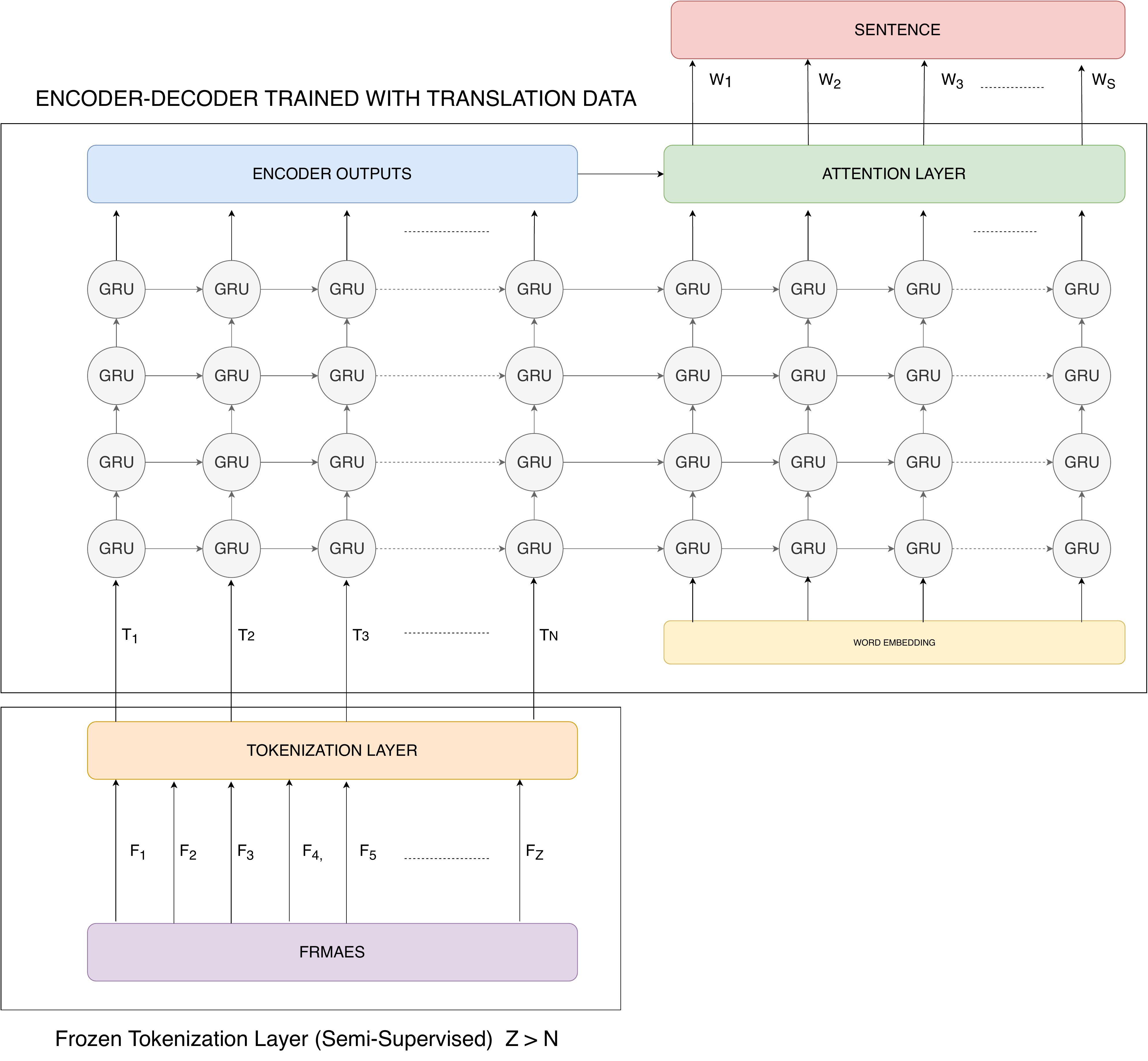}
\caption[b]{NSLT with Semi-Supervised Tokenization}
\label{NSLT}
\end{figure}
	
\paragraph{End-to-end Frame Level Tokenization}
	End-to-end learning of all parts of the system, namely tokenization and sequence-to-sequence, is considered as one of the breakthroughs in DL. It gives state-of-the-arts results in the presence of sufficiently large data in many domains. Additionally, it diminishes the burden of feature engineering along with model interpretability. \cite{camgoz2018neural} has no explicit tokenization and does not give insights on what the tokenization layer learns. Alexnet architecture pre-trained on ImageNet \cite{alexnet} is deployed to extract non-linear frame level spatial embeddings.  To test the validity of this approach, we try feeding different body parts, such as hands, to the system as inputs. This is expected to test the capability of the end-to-end framework to attend to significant parts of images. We know that hand shapes of a signer convey crucial information.  Hence, we prepare three settings with the same system parameters suggested by \cite{camgoz2018neural}, but with different inputs: First input type is the full frame; second is right hand crops and the last one is the concatenation of right hands and left hands side by side. Note that we ignore left-hand setting alone for simplicity as we observe that all the signers are right handed. 
\paragraph{Tokenization with Keypoints of Body Parts} 
	One of the  problems with using keypoints is background noise and different camera setups in datasets. It is believed that OpenPose is very robust to these type of challenges and is able to provide accurate keypoint locations. However, motion blur leads to noise in hand keypoints. Our prior knowledge indicates that hand shapes have high intra-class and low inter-class variations. Regarding this, the noise level in OpenPose hand keypoints hinders the efforts to achieve efficient tokenizations for SLT. In the translation dataset  \cite{camgoz2018neural}, all images contain one person with no occlusions. Therefore, it is simple to identify hands and body poses. In this work, we are able to evaluate the performance of it by comparing other methods in hand shape identification. Besides that, the trajectory of body parts in videos will be clearly traced by OpenPose. This might be fruitful for tokenization. However, there is no golden rule on how to convert keypoints into good representations for NSLT. The study in \cite{korean} attaches special attention to this problem and proposes a method called "Object 2D Normalization". We adopt it into our experiments. To provide a consistent comparison setup with other tokenization approaches, we again use three different experimental setups with the same network parameters. Those are right hand keypoints, both hands keypoints and full body pose, ignoring face keypoints.
\paragraph{Tokenization with 2D-CNN trained on predetermined task}
	We deploy a 2D-CNN trained for hand-shape
	classification using different datasets for tokenization. We further use multitask learning and domain adaptation as detailed in Section IV. As opposed to \cite{camgoz2018neural}, we freeze the CNN and do not tune it  while training the sequence-to-sequence model. This enables us to observe the effects of different methods. It may be possible to attain higher scores compared to end-to-end training by enforcing additional supervision, if data is available.  We seek for different methods to transfer this sort of supervision to the NSLT system. We also investigate domain adaptation and multitask learning for different cases to examine the effectiveness.
	    \par First, we analyze the transfer capability of the resulting system. We trained our CNN with data coming from different domains. There is no strong relationship between sign languages of different communities. Experiments would illustrate whether a transfer is possible between different sign languages. Secondly, we have two different datasets labeled with different approaches. One of them is weakly annotated according to hand shapes predefined by linguists and contains considerable noise and blur. The second is labelled with data driven hand shapes determined by semi-supervision. Lastly, we are able to compare the contributions of having labels from the target domain with side information. The state-of-the-art approach, sign-to-gloss-to-text, proposed in \cite{camgoz2018neural} utilizes gloss level supervision from the target domain. As a result, we can compare the actual improvement of our method against this sign-to-gloss-to-text.
\paragraph{Tokenization with 3D-CNN Features Trained for Action Recognition}
Contrary to 2D-CNNs, 3D-CNNs also convolve in time and  extract shorter feature sequences from videos. This feature seems to be advantageous as it decreases the burden of long term dependencies. Sign language videos also include  lots of redundant frames. However, action recognition is a different task and it is doubtful that the representations obtained by it are suitable to feed  the sequence-to-sequence scheme. 3D-CNNs model sequences  implicitly to some level. Therefore, we include  3D-CNNs trained on Action Recognition datasets for comparison  with other tokenization methods, and test whether the representations learned from actions are useful to represent signs.

\section{Methodology}
\subsection{Datasets}

	For translation experiments, we use  RWTHPHOENIX-Weather 2014T introduced by Camgoz et al. \cite{camgoz2018neural}. It is an extended version of RWTHPHOENIX-Weather 2014 which is a SLR benchmark. This dataset is the only publicly available challenging dataset to the best of our knowledge. To conduct different experimental
	setups, all data has been processed with Openpose \cite{cao2018openpose} to extract body and hand keypoints and locate the hands in the frames. As a result, we obtain three different settings, image crops including only the right hand, both hands, and full frames. We need the right hand since signers are right-handed. The dataset has 7096 training videos with varying length sequences. The size is sufficient to train a deep model, with predefined 642 test videos and 519 dev videos.
	\par For CNN training, we used the dataset consisting of over one million images with weak annotations collected from three different sources prepared by Koller et al. \cite{koller16:deephand}. Those sources are Danish \cite{danish} and New Zealand \cite{nz} SL Dictionaries along with RWTHPHOENIX-Weather 2014 \cite{forster2} labeled for 61 context independent hand shapes including one junk class. As mentioned in \cite{camgoz2018neural}, RWTHPHOENIX-Weather 2014 training set does not overlap with the test set of the translation data, RWTHPHOENIX-Weather 2014T. The test set of RWTHPHOENIX-Weather 2014 consists of 3359 challenging frames labelled manually for hand shapes. Although it is safe to use all the data, we spare RWTHPHOENIX-Weather 2014 in some experiments for deeper analysis. There is an independent dataset \cite{doga} from another domain, respectively small in size and noise level compared to the previous one. It has 56100 frames, 30 signs and seven signers collected in an isolated environment. It has a different annotation methodology and context dependent labeling: It covers a limited number of hand shapes; but annotations refer to the context of the hand, such as the body part the hand is pointing to, as well as the hand shape. Another key difference is that it is collected in a semi-unsupervised manner with 45 different class labels without a junk class. We use this dataset for domain adaptation and multi-task learning experiments.
\subsection{2D-CNN Training Methods }
	In this section, we discuss several computer vision methods that might solve specific challenges in hand shape classification. To begin with, data is generally weakly labeled, as mentioned earlier. Additionally, backgrounds and environments are distinctly different between the two datasets. This prevents 2D-CNNs from being robust to different conditions. This occurs if there is data from the target domain. Despite that, we consider methods to facilitate resulting representations by utilizing available side information.  
\paragraph{Using Sequence Information to Refine Lables} 
	Koller et al. \cite{koller16:deephand} proposed a CNN-HMM hybrid method that uses sequence information to eliminate noise in annotation. While a CNN learns to identify hand shapes, a HMM learns to correct labels jointly. This enables higher scores in SLR, benefiting from the huge amount of data collected.  In sign language  videos, there is a natural temporal ordering. We know that consecutive frames have similar shapes and a temporal model could be established to refine wrong labels. The sequence of hand shapes are expected to follow a certain probabilistic pattern. It also enables the CNN to encapsulate additional information meanwhile it could harm generalization of the network. 
\paragraph{Multitask Learning}
	Multitask learning learns better image representations by training on related tasks jointly. In our case, the two tasks are defined in two different datasets as seen in Fig. 2. The first dataset contains one million images involving considerable amount of image blur and noise in labels. Meanwhile,  the  second dataset,\cite{doga} is respectively small and less noisy. The class definitions of the two datasets are different. Hence, the annotations differ semantically and functionally. The latter dataset covers limited hand shapes, but annotations convey more contextual information. Therefore, the CNN can learn  the interaction between the hand and the  body parts in the background. However, it needs some effort to find a optimal schedule to train them jointly. As our network is optimized considering two objectives jointly, a balancing mechanism between tasks is needed.

\begin{figure}[b]
\centering
\includegraphics[width=8cm]{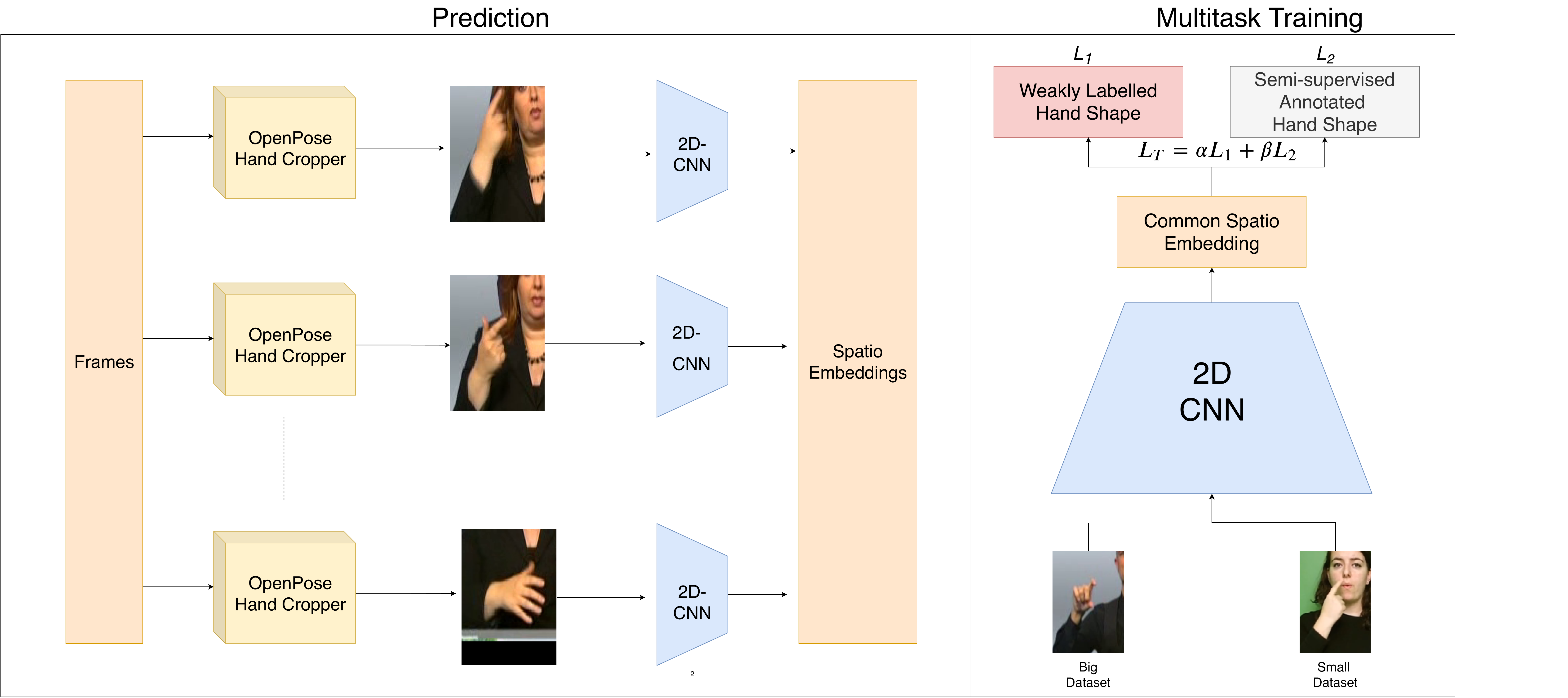}
\caption[b]{Multitask Tokenization Method}
\label{MT}
\end{figure}

\paragraph{Domain Adaptation}
	It is observed that 2D-CNNs are not effective for data coming from a domain different than the source domain. Ganin et al. \cite{gradientreversal} proposed an adversarial method, a simple idea that prevents 2D-CNNs from learning domain specific features by adding a gradient reversal layer. It is expected to increase classification scores, but we look for better representations  in tokenization. This might show whether NSLT models can handle domain shifts implicitly or an explicit effort is required.

\subsection{Sequence-to-sequence Model }
    In this study, we use existing successful sequence-to-sequence models in our experiments. Transformers \cite{transformer} have dominated NLP tasks. Hence, we include them in our experiments along with other popular LSTM-based methods. We also try successful attention mechanisms such as Luong Attention\cite{luong-etal-2015-effective}, and Bahdanau Attention \cite{bahdanau}. The formalized structure by \cite{camgoz2018neural} is suitable to adapt to the new methods. It is clear that an attention mechanism is need as our videos consist of long frame sequences and no reduction technique is applied.

\section{Experiments}
	For end-to-end experiments, we use Luong Attention and the same hyper-parameters provided by \cite{camgoz2018neural} with different inputs, namely, right hand, both hands and full frame. To conduct other experiments, we use OpenNMT framework \cite{opennmt} which enables  the implementation of complex models easily with various configurations. To evaluate our translation results, we used the same metrics in \cite{camgoz2018neural}, BLEU (BLEU-1, BLEU-2, BLEU-3, BLEU-4) \cite{bleu} and ROUGE (ROUGE-L F1-Score) \cite{rouge}, to sustain consistent comparisons. 
	\par For CNN training, we select Inception-v3 \cite{inception}  architecture with pre-training on ImageNet for all experiments. As our hand shape datasets are heavily class imbalanced, we deploy a mini-batch sampler to  balance the class representations. Adam \cite{adam}  optimizer is used   for all experiments except domain adaptation as they converge with SGD with momentum instead of Adam. To perform 3D-CNN experiments, we use state-of-the-art RGB I3D model \cite{I3D} trained on the Kinetics Dataset.

\subsection{Input Analysis of End-to-end Learning}

    To conduct an input analysis, we replicate the end-to-end scheme experiments with predefined input types (Table~\ref{table:endtoend}). The full frame results are directly copied from the sign-to-text results with Luong Attention in \cite{camgoz2018neural}. The columns give the ROUGE and BLEU scores. We base our comparisons on BLEU-4 scores, although other scores also indicate the same results. The best results are indicated in bold. Using the full frame, a BLEU-4 score of 9.0 is obtained. We observe that although using the right hand performs better than the full frame (10.25), the best results are obtained by inputting both hands: 11.66, a significant improvement. However, when we direct our attention to the tokenization experiments in Table~\ref{table:wotarget}, we observe that tokenization enables the network to focus on important body parts, as discussed below. In those experiments, full frames perform better except for a singular case.

\begin{table}[b]
\begin{center}
\caption{Test Results of End-to-end scheme }
\begin{tabular}{|p{1.4cm}||p{1cm}|p{1cm}|p{1cm}|p{1cm}|p{1cm}|}
 \hline
  Input & ROUGE & BLEU-1 &  BLEU-2&  BLEU-3&  BLEU-4 \\
 \hline
 Full  &   30.70  & 29.86   &  17.52  &  11.96   &  9.00  \\
 \hline
 Both Hands &  \bftab  34.53 &   \bftab  33.65  &  \bftab  21.01 &   \bftab  15.19 &   \bftab  11.66  \\
 \hline
 
 Right Hand & 31.89 & 30.57  &  18.67 &  13.19 & 10.25   \\
 \hline
\end{tabular}
\label{table:endtoend}
\end{center}
\end{table}
\subsection{Architecture \& Attention Mechanism Search}
 \par For search we use the same hyperparameters as in Camgoz et al. \cite{camgoz2018neural}, except batch size and hidden unit size and perform tuning for only Transformer models having a very different architecture from Bahdanau and Luong Attention mechanisms. We choose batch size of 16.
 We choose tokenization with a frozen 2D-CNN trained on Imagenet with full frames as input to compare and analyze different architectures. Table~\ref{table:arch} gives the comparison of different attention mechanisms. Consistent with earlier research, all models converge to a local optimum and perform well, but Bahdanau outperforms others as seen in row 2 of Table~\ref{table:arch}. To our knowledge, this is the first study training a Transformer for this dataset and the results are  better than Luong Attenntion. We believe that Transformer architectures are more suitable for bigger datasets and the results are consistent with results of earlier studies in this dataset. We choose Bahdanau attention to report further experiment results.
 
\begin{table}[t]
\caption{Comparison of  Attention Mechanisms }
\begin{tabular}[t]{|p{1.3cm}||p{1cm}|p{1cm}|p{1cm}|p{1cm}|p{1cm}|}
 \hline
  Model & ROUGE & BLEU-1 &  BLEU-2&  BLEU-3&  BLEU-4 \\
 \hline
  Bahdanau & \bftab  29.41 & \bftab  30.46  &  \bftab 17.79 &  \bftab 12.36 &  \bftab 9.40  \\
 \hline
 Transformer  & 28.20 & 28.66  & 16.33 &  11.15 &  8.43  \\
 \hline
  Luong  & 26.94 & 27.46  &  15.12  &  10.38  &  7.93  \\
 \hline
\end{tabular}
\label{table:arch}
\end{table}

\begin{table*}[t]
\centering
\caption{Test Results of Experiments without Access to Target Domain Annotations}
\begin{tabular}{|p{5cm}||p{1cm}|p{1cm}|p{1cm}|p{1cm}|p{1cm}|}

 \hline
  Tokenization Approaches & ROUGE & BLEU-1 &  BLEU-2&  BLEU-3&  BLEU-4 \\
 \hline
 \multicolumn{6}{|l|}{ \bftab  Group 1}  \\ 
 \hline
 Pre-Trained 2D-CNN with Full Frames & 29.41 &   \bftab 30.46  & \bftab   17.79 &  \bftab   12.36 &  \bftab  9.40   \\
 \hline
 
 Pre-Trained 2D-CNN with Both Hands &  \bftab  29.62 & 30.24 & 17.46  &  12.14&  9.33  \\
 \hline
 
 Pre-Trained 2D-CNN with Right Hands & 29.09 & 29.51  &  17.03 &  11.79 & 9.06   \\
 \hline
  \multicolumn{6}{|l|}{ \bftab Group 2}  \\ 
 \hline

 Pre-Trained I3D  with Full Frames  &  \bftab  29.74 &  \bftab  29.52  &   \bftab  17.09 &   \bftab  11.64 &  \bftab  8.76   \\
 \hline
 
  Pre-Trained I3D  with Both Hands & 28.64 & 28.27 &16.14  &  10.99&  8.26  \\
 \hline
 
  Pre-Trained I3D  with Right Hands & 28.00 & 28.01  &  15.78 &  10.73 & 8.09   \\
 \hline
  \multicolumn{6}{|l|}{\bftab Group 3}  \\ 
 \hline
  
 Keypoints of Body \& Both Hands   & \bftab 32.85 & \bftab 33.18  & \bftab 20.39 &  \bftab 14.26 &  \bftab 10.92  \\
 \hline
 Keypoints of Both Hands  & 31.47 & 31.51  &  19.08 &  13.26  & 10.23  \\
 \hline
  KeyPoints of Right Hands & 30.65  &  30.69  & 18.49 & 12.89 &  9.91  \\
 \hline
 
 \multicolumn{6}{|l|}{\bftab  Group 4}  \\ 
  \hline

  CNN trained on Big Dataset  & 34.59 & 35.52  &  22.37 &  15.80 &  12.17  \\
 \hline
  CNN trained on Small Dataset & 31.98 & 33.40  &  20.53 &  14.50  & 11.15  \\
 \hline
 Domain Adaptation & 34.41 & 34.84  & 22.07 & 15.75 &  12.21  \\
 \hline
 Multitask  & \bftab 36.28 & \bftab 37.22  &   \bftab 23.88 &   \bftab  17.08 &  \bftab 13.25   \\
 \hline
\end{tabular}
\label{table:wotarget}
\end{table*}

\subsection{Tokenization Independent of Target Domain}

        In this section, we focus on alternative tokenization methods ignoring knowledge from the target domain. Therefore, we do not tune the tokenization methods with the sequence-to-sequence model. OpenPose outputs, 3D-CNNs and 2D-CNNs are deployed to the tokenization layer in those experiments. We also apply Multitask learning and domain adaptation to train our 2D-CNNs to increase generalization of video representations.
\paragraph{Effects of a Frozen 2D-CNN trained on ImageNet }
    Group 1 in Table~\ref{table:wotarget} indicates the results of this tokenization method. The full frame setting gives the best performance, 9.40 in BLEU-4 scores. Both hand setting outperforms others according to the ROUGE score. However, there is no significant difference between input types. ImageNet is diverse and huge. Those properties of the dataset enables knowledge transfer to the sign language domain. Consulting  Table~\ref{table:endtoend}, we observe that the CNN pre-trained on ImageNet in the end-to-end scheme cannot be tuned well in a full frame setting as the frozen CNN provides comparable results listed in the first row of Group 1 of Table~\ref{table:wotarget} compared to the first row of Table~\ref{table:endtoend}. Although they are both trained on ImageNet, the architectures are different. The performance of the end-to-end model tends to increase dramatically by feeding with the hand crops whereas results of the frozen network increases with those crops. 

\paragraph{Effects of Temporal Convolution}
    The experiment results related to 3D models, which represent the temporal nature are illustrated in Group 2 of Table~\ref{table:wotarget} with different input types. Similar to other approaches, I3D model \cite{I3D}, an adapted version of Inception \cite{incep} to action recognition domain, is the most successful in the full frame setting, obtaining 29.75 ROUGE and 8.76 BLEU-4 scores. Those incremental results, from using right hands to full frames are consistent with the results of approaches listed in Group 1-3 of Table~\ref{table:wotarget}. We infer that generic models trained on huge dataset perform well in full frame as they are robust to different type of noises. We should compare this results with 2D-CNNs to gain two insights. The frozen 2D-CNNs are better than 3D-CNNs in terms of BLEU-4 score, but worse in ROUGE score. Hence, it is not possible to claim a superiority between them. However, Convolution in time reduces features in length for sequence modeling and this contributes to the acceleration of convergence. Furthermore, our experiments  on sentence level and 3D-CNNs are useful for modeling longer sequences than sentences. We leave the search for optimal combination of 3D-CNNs and sequence-to-sequence models as future work.
    
\paragraph{Capacity of Keypoints to Represent Signs }
    We report the results of those experiments in Group 3 of Table~\ref{table:wotarget}. We use hand keypoints and body poses to sustain consistent comparison. Again, using all body parts ignoring faces will give the highest score in all metrics and a significant leap occurs by adding location information of hands in space. Also, both hands are more useful than the right hand alone as the performance increases over 0.5 points in all metrics. Keypoints are very refined representations of body parts and the information is enforced explicitly. The experiments with key points show clearly the importance of interaction between body parts. However, they are not good enough to replace with CNNs trained for hand classification as seen in Group 3 and Group 4 of Table~\ref{table:wotarget}. Features coming from the last layer of the hand shape classifiers outperforms key points in all metrics and it shows that key points cannot identify the sign language hand shapes as well as CNNs for translation if we look at the third row of Group 3 and the second row of Group 4.

    In this section, we aim to have good and generic representations of hand shapes. Hence, we conduct remaining experiments with only right hands. We use two different datasets. The first dataset is a big dataset containing nearly one third of the hand images collected in \cite{koller16:deephand}. The collected dataset contains over one million hand images and we exclude RWTHPHOENIX-Weather 2014 dataset for the experiments in this section. Because, It is from the same subject domain, weather report, as the translation dataset of \cite{camgoz2018neural} and there is overlap between the two datasets.  The second dataset is small \cite{doga}. We name the first one as Big and the second one as Small in Table~\ref{table:wotarget}.
    \par We first check whether the networks are sufficiently trained by inspecting classification accuracy.  In Table~\ref{table:cnn}, the classification performances of our 2D-CNNs are provided for detailed analysis. The first rows shows baseline results. Data augmentation with pre-processing means that we add random noise in image dimensions of brightness, contrast, saturation, hue. We observe that data augmentation shows a very small improvement. Multitask shows the result of joint training with the Small and Big datasets; with 10 \% weight for the small task's objective. Domain adaptation refers to the network with gradient reversal layer to accomplish a domain shift toward the translation dataset. Our metrics are Top-1 and Top-5 Accuracy. The table shows that the Domain Adaptation method raises Top-1 accuracy over 10 points and this is a strong clue of domain difference. Multitask learning achieves 76.77 Top-1 and 92.88 Top-5 accuracy.  We understand the Domain Adaptation network has accomplished to provide domain dependent features. There might be two underlying factors why Multitask learning performs almost as good as Domain Adaptation. The first one, learning from two different domains can benefit a third domain. Secondly, the labeling difference in the Small dataset may result in more meaningful representations.
    \paragraph{Effectiveness of Hand Shapes}
\begin{table}[b]
\begin{center}
\caption{Classification Results on Target Domain}
\begin{tabular}{|p{3cm}||p{1cm}||p{1cm}|}
 \hline
  Methods &  Top-1 &  Top-5   \\
  \hline
  Baseline  &  65.49  &  89.49    \\
 \hline
  Data Augmentation  &  67.07  &  90.65   \\
 \hline
  Multitask &  76.77  &  92.88   \\
 \hline
  Domain Adaptation  &  78.74  &  94.73   \\
 \hline
\end{tabular}
\label{table:cnn}
\end{center}
\end{table}
    \par Group 4 in Table~\ref{table:wotarget} illustrates the success of tokenization with trained CNNs on hand shape classification. In first row of Group 4, We see a significant improvement with the multitask approach that increases the base score from 12.17 to 13.25 in BLEU-4 along with a rise in all metrics. However, Domain Adaptation does not improve translation quality and even leads to a slight decline in all metrics expect BLEU-4. Besides the Multitask network, a simple network trained on hand shape classification performs better than other methods introduced earlier in Group 1-3 of Table~\ref{table:wotarget}. First, We observe that hand shapes play a critical role in sign translation. Second, it is possible to transfer knowledge between different sign languages. Furthermore, domain difference is not a big problem for sequence-to-sequence models as Domain Adaptation does not improve results. It implies that sequence-to-sequence models can handle the domain shift implicitly while training. We show that a simple network trained on a different but related task can be deployed to the tokenization layer regardless of its training domain. Thanks to knowledge transfer, improved hand shape representations regardless of domains result in better translation.

\subsection{Utilizing Information From Target Domain}

\begin{table}[b]
\begin{center}
\caption{Test Results with Target Annotations}
\begin{tabular}{|p{1.3cm}||p{0.8cm}|p{1cm}|p{1cm}|p{1cm}|p{1cm}|}
 \hline
  Tokenization & ROUGE & BLEU-1 &  BLEU-2&  BLEU-3&  BLEU-4 \\
 \hline
  DeepHand & \bftab  38.05 & \bftab  38.50  &  \bftab 25.64 &  \bftab 18.59 &  \bftab 14.56  \\
  \hline
  Multitask & 36.35 & 37.11  &  24.10  &  17.46 &  13.50  \\
 \hline
  Baseline  & 35.22 & 35.97  & 23.10 &  16.59 &  12.89  \\
 \hline
\end{tabular}
\label{table:target}
\end{center}
\end{table}

    In this experiment, we use all the datatests to see the amount of improvement and how close the proposed tokenization methods are to the state-of-the-art method. RWTHPHOENIX-Weather 2014 dataset, not utilized to date, has about two times more frames than all other datasets combined. As a consequence, the resulting dataset consists of over one million frames. The results are listed in Table~\ref{table:target}. DeepHand refers to the network trained in \cite{koller16:deephand} and Multitask refers to the network jointly trained on over one million images with Small dataset as in the previous section. Baseline refers to the network trained on over one million images with weak annotations. Multitask  in Group 4 of Table~\ref{table:wotarget} is higher than the results of the Baseline in Table~\ref{table:target}. It suggests that Multitask learning with less data is better than training on huge amount of weakly labeled data for tokenization. DeepHand using sequence information to refine frame labels achieves the best results, 14.56 BLEU-4 and 38.05 ROUGE scores where Multitask learning achieves 13.50 BLEU-4 and 36.35 ROUGE scores. The state-of-the-art model in \cite{camgoz2018neural} first extracts glosses then translates glosses to texts, called S2G2T model achieves 18.13 BLEU-4 and 43.80 ROUGE scores. S2G2T performs direct translation from frames, but gloss level annotations are required for training. Also, it benefits from full frames while our tokenization focuses on only the right hands. Group 1-3 of Table~\ref{table:wotarget} indicates inputting full frames gives higher scores. Note that both of the methods, S2G2T and ours are initially trained with target domain data. In short, this comparison shows that we closed the gap between methods requiring gloss annotations and sign-to-text translation without explicit gloss representations.

\section{Conclusion}
    We investigated the use of tokenization learning for sign to text translation. If gloss level annotations are available, these can be utilized for achieving better tokens. However, these annotations are costly and data annotated with glosses is scarce. We instead use other sign datasets to learn tokenization. Since the datasets are from different domains, we try domain adaptation and multitask learning for better tokenization. We use both 2D and 3D architectures. We have shown that 3D-CNNs would be efficiently used for SLT in the future and have enabled knowledge transfer between Sign Languages. Furthermore, we illustrated contributions of different body parts to the quality of translations in many tokenization approaches. Eventually, our approaches were compared with the state-of-the-art model to show that laborious gloss annotation might not be needed to achieve higher scores in the future.
    \par For future work, a new method combining right hand information with the context of a frame may be found to exceed performance of the state-of-the-art method.  Secondly, Sign Languages are highly dependent on whole the context instead of sentence parts. Hence, 3D-CNNs could be used to summarize long sequences thanks to spatio-temporal filters.


\bibliographystyle{plain}

\bibliography{main}

\end{document}